\def\year{2022}\relax
\documentclass[letterpaper]{article} 
\usepackage{aaai22}  
\usepackage{times}  
\usepackage{helvet}  
\usepackage{courier}  
\usepackage[hyphens]{url}  
\usepackage{graphicx} 
\usepackage{natbib}  
\usepackage{caption} 
\DeclareCaptionStyle{ruled}{labelfont=normalfont,labelsep=colon,strut=off} 
\usepackage{algorithm}

\usepackage{newfloat}
\usepackage{listings}
\floatstyle{ruled}
\newfloat{listing}{tb}{lst}{}
\floatname{listing}{Listing}

\usepackage{subcaption}
\usepackage{algorithm}
\usepackage{algpseudocode}
\usepackage{multirow}
\usepackage{booktabs}
\usepackage{blindtext}
\usepackage{tikz}
\usetikzlibrary{decorations.pathreplacing,calc}
\usepackage[switch]{lineno}
\renewcommand{\paragraph}[1]{\vspace{1mm}\noindent\textbf{#1}}

\usepackage{kotex}
\title{Iteratively Selecting an Easy Reference Frame \\Makes Unsupervised Video Object Segmentation Easier}
\author {
    Youngjo Lee,
    Hongje Seong,
    Euntai Kim\footnote{Corresponding author.}
}
\affiliations {
    School of Electrical and Electronic Engineering, Yonsei University, Seoul, Korea\\
    \{lzozo95, hjseong, etkim\}@yonsei.ac.kr
}

\usepackage{bibentry}

\begin{document}


\maketitle

\begin{abstract}
Unsupervised video object segmentation (UVOS) is a per-pixel binary labeling problem which aims at separating the foreground object from the background in the video without using the ground truth (GT) mask of the foreground object. Most of the previous UVOS models use the first frame or the entire video as a reference frame to specify the mask of the foreground object. Our question is why the first frame should be selected as a reference frame or why the entire video should be used to specify the mask. We believe that we can select a better reference frame to achieve the better UVOS performance than using only the first frame or the entire video as a reference frame. In our paper, we propose Easy Frame Selector (EFS). The EFS enables us to select an ``easy'' reference frame that makes the subsequent VOS become easy, thereby improving the VOS performance. Furthermore, we propose a new framework named as Iterative Mask Prediction (IMP). In the framework, we repeat applying EFS to the given video and selecting an ``easier'' reference frame from the video than the previous iteration, increasing the VOS performance incrementally. The IMP consists of EFS, Bi-directional Mask Prediction (BMP), and Temporal Information Updating (TIU). From the proposed framework, we achieve state-of-the-art performance in three UVOS benchmark sets: DAVIS16, FBMS, and SegTrack-V2.
\end{abstract}

\section{Introduction}

\noindent Video Object Segmentation (VOS) is a task that segments the objects as pixel-level binary masks from videos. In general, VOS is divided into two other tasks: Semi-supervised Video Object Segmentation (SVOS) and Unsupervised Video Object Segmentation (UVOS). At train time, video frames and their ground truth (GT) masks are given to both tasks. They are trained in a supervised manner. 
At inference time, the first frame’s mask annotation of foreground objects is given to SVOS. On the other hand, only the video frames without any annotations are given to UVOS. In this paper, we propose a novel framework for UVOS. 

In SVOS, as the first frame’s GT mask is given, most SVOS models indeed predict the remaining frames’ masks with the first frame as a reference frame. 
For UVOS, there is no need to use the first frame as a reference frame because there is no annotation even for the first frame. Thus, any frame can be a reference frame in UVOS. Especially, we call the reference frame an easy/hard frame according to how easy to predict a foreground object for the remaining frames from it.

\begin{figure}[t]
\centering
\includegraphics[width=\linewidth]{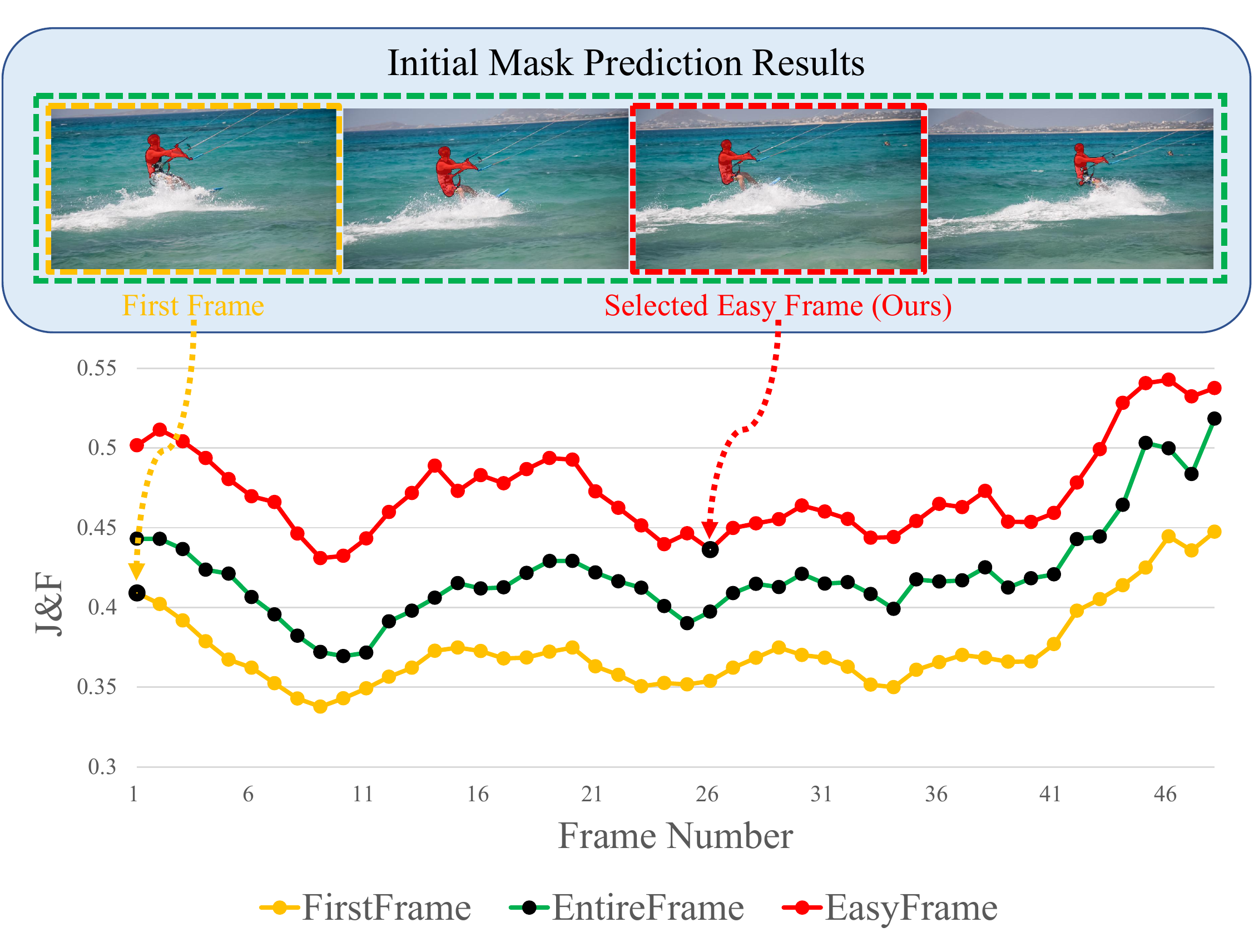}
\caption{Three experimental results according to the reference frame selection: (1) referencing the first frame only (yellow), (2) referencing entire frames (green), and (3) referencing the easy frame which predicted by our Easy Frame Selector (red).
We plot the $\mathcal{J}$\&$\mathcal{F}$ scores computed by predicted mask for every frame according to the selection. Black dots indicate the selected reference frame(s).
Our approach (red) shows superior performance than other reference frame selection methods (yellow and green) in every frame. Note that predicted masks are express on the image and the dotted arrows indicate the positions of each frame.
}
\label{figure1}
\end{figure}

Most UVOS methods \citep{tokmakov2017learning,jain2017fusionseg,tokmakov2017learning2,cheng2017segflow,song2018pyramid,yang2019anchor,lu2019see,wang2019learning,faisal2019exploiting,siam2019video,tokmakov2019learning,zhou2020motion,zhuo2019unsupervised} find the object’s mask based on various types of reference frames. In \citet{yang2019anchor}, they call the first frame as an anchor frame and use it as a reference frame. However, the first frame is not always the best reference frame as shown in Figure \ref{figure1}. To enlarge information to refer, some papers \citep{tokmakov2017learning,song2018pyramid,lu2019see,wang2019learning,faisal2019exploiting,tokmakov2019learning,zhuo2019unsupervised} use the larger number of reference frames.  Especially, \citet{song2018pyramid,wang2019learning,faisal2019exploiting} use entire frames as reference frames.
In this case, the chance to use easy frames as reference frames is enhanced. But, the chance to use hard frames as reference frames is also enhanced at the same time. Hard frames can deteriorate the good prediction from easy frames and can lead to worse performance. 

In this paper, to address the above issues, we propose Easy Frame Selector (EFS), which can select easy frames. In Figure \ref{figure1}, the yellow line shows the accuracy of mask prediction using the first frame as a reference frame. The accuracy is evaluated with $\mathcal{J}$\&$\mathcal{F}$ which is an evaluation metric for DAVIS. Here, when we use the entire frames as reference frames to increase the number of the reference frame as like in \citet{song2018pyramid,wang2019learning,faisal2019exploiting}, the accuracy is slightly raised, as shown in the green line. 
The red line shows that we surpass the other two even with a single reference frame selected by EFS. It definitely shows why EFS is needed.

EFS takes a pair of an image and its predicted mask acquired from Salient Object Detection (SOD) as an input in a frame-by-frame manner. Then, EFS estimates the difficulties of the frames and makes a top-k selection as reference frames based on the difficulties. Due to EFS, we can prevent the propagation of inaccurate masks by filtering out the hard frames.

After selecting easy frames, we predict masks for the remaining frames from easy frames. 
Then, the quality of predicted masks becomes better than the masks acquired by SOD. And, this better-quality masks can be used as saliency cues recursively. Again, we select new easy frames from the better-quality masks. Then, we predict masks for the remaining frames from the newly selected easy frames. This iterative process is repeated and we can get accurate masks. We propose this iterative structure as Iterative Mask Prediction (IMP), which makes an easy frame be easier and can predict better-quality masks from the easier frame. 

In this paper, we focus on selecting easy frames to be used as reference frames. We show that using easy frames as reference frames is how effective by many experiments. Also, We update the saliency cues to get richer temporal information by our novel iterative structure. Hence, easy frames gradually become easier so that our proposed framework improve the accuracy by a large gap. Our main contributions can be summarized as follows:

\begin{itemize}
\item We propose Easy Frame Selector module that selects the top-k easy frames. 
\item We propose Iterative Mask Propagation which can make a better-quality prediction from the easier frame and far easier frame selection from the better-quality masks iteratively. 
\item We achieve state-of-the-art performance at three UVOS benchmark sets (DAVIS16, FBMS, and SegTrack-V2) from our simple yet powerful framework.
\end{itemize}

\begin{figure*}[!h]
\centering
\includegraphics[width=\linewidth]{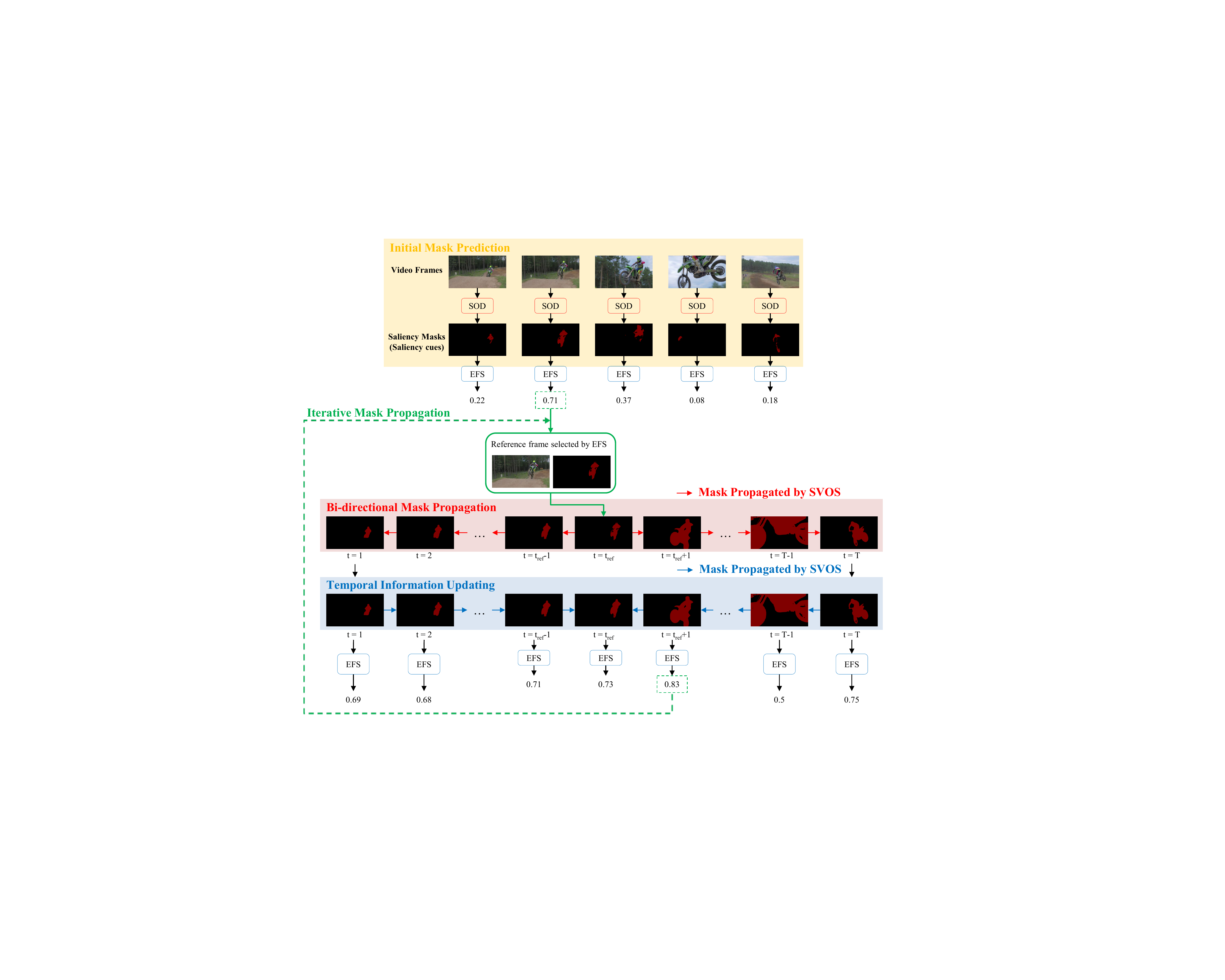}
\caption{Overall architecture for our proposed method. First, each video frame is fed to SOD network (empty red block) to get their saliency masks. This process is Initial Mask Prediction (yellow-filled-bloack). EFS (empty blue block) estimates their qualities. The numbers connected by downside arrows from EFS are the estimated scores of the pairs of images and masks. Then, EFS selects easy frame(s) with the highest score as a reference frame. (For straightforward depiction, we only show a case of using a single easy frame.) From the reference frame, the mask is propagated by SVOS bi-directionally (red arrows). By propagating the masks at both ends to the selected easy frame, saliency cues are updated with temporal information (blue arrows). Then, we select the new easy frame(s) with EFS again. These processes are repeated by Iterative Mask Propagation (IMP) (green dotted line). At the final iteration, the entire process ends with Bi-directional Mask Propagation (BMP) (red-filled-block).}
\label{figure2}
\end{figure*}

\section{Related Work}

\subsection{Semi-supervised Video Object Segmentation}

Video Object Segmentation (VOS) represents the tasks that segment the object’s mask in a video sequence. Semi-supervised Video Object Segmentation (SVOS), one of the subcategories of VOS, is a task that provides the object’s mask in the first frame as a clue to segment. In other words, it predicts the foreground’s mask with the first frame as a reference frame.

In \citet{griffin2019bubblenets}, they raise doubts about it. Although SVOS task provides the GT mask at the first frame, they experiment to provide the mask at another frame. They observe that some other frames make a better performance than the first frame as a reference frame.

We also think that the selection of a reference frame is important. We have a similar motive, but our solution is different because of the disparity between the tasks. They focus on SVOS, so they aim to determine which frame to annotate instead of the first frame. On the other hand, we cannot have any annotation for every frame in UVOS, but it is still important to have a good reference frame. Therefore, we propose Easy Frame Selector (EFS), which can select the easy frame. 

\subsection{Unsupervised Video Object Segmentation}

Unlike SVOS, UVOS does not provide GT annotation for every frame in videos at inference time. 
Since only the video, which is a series of RGB images, is given for UVOS, there is no given information to refer to like in SVOS. Thus, we need to find such information to refer to in a video. This useful information is called saliency cues. 

In \citet{jain2017fusionseg,wang2019learning,faisal2019exploiting,tokmakov2019learning,zhuo2019unsupervised}, they use the object saliency cue. Object saliency cue is a  localization of a frame-level object, through General Object Detector (GOD), Salient Object Detector (SOD) or other basic object detectors such as Faster-RCNN \citep{ren2016faster} and YOLO \citep{redmon2016you}. It tells where the foreground object exists. It is often used by combining it with motion saliency cues to predict the masks in some papers \citep{jain2017fusionseg,faisal2019exploiting,tokmakov2019learning,zhuo2019unsupervised}. Also, it is also used to predict adding some temporal information by LSTM in \citet{wang2019learning}.

In this paper, we also exploit the object saliency cue. However, the object saliency cue has a fatal disadvantage. It does not contain any temporal information. Due to the lack of temporal information, it is difficult to discriminate similar instances. To overcome the disadvantage, we propose Iterative Mask Propagation (IMP). IMP transfers temporal information to the saliency cues by propagation. By iteratively updating the saliency cues with temporal information, similar instances can be discriminated by acquired temporal information. 

\subsection{Salient Object Detection}

Salient Object Detection (SOD) is a task of finding the most salient object in an image. Recently, many papers \citep{wu2019stacked,liu2018picanet,huang201950,wu2019mutual,wang2020progressive} get good performances by extracting various features from an image, from fine features to coarse features. 
Among them, we adopt PFPN \citep{wang2020progressive} as our SOD model to extract a saliency cue from each image. PFPN \citep{wang2020progressive} fuses various features in depth-level to enhance the features, so it can detect salient objects with various scales. Due to their ability to detect multi-scale objects, we adopt PFPN \citep{wang2020progressive} as our SOD model.

\section{Proposed Method}

\begin{table*}[t]
\centering
\caption{Quantitative results on DAVIS16 validation set. The best results for each metric are \textbf{bold faced}. All the other results are borrowed from \citet{lu2020video,zhen2020learning,liu2020f2net}.}
\label{table1}
\resizebox{0.95\linewidth}{!}{
\begin{tabular}{@{}ccc|ccccccccccc@{}}
\toprule
\multicolumn{3}{c|}{Model}                           & MSG      & NLC  & CUT  & FST    & SFL  & LMP    & FSEG   & LVO   & UOVOS & ARP   & PDB  \\ \midrule
\multirow{3}{*}{$\mathcal{J}$} & Mean   & $\uparrow$   & 53.3     & 55.1 & 55.2 & 55.8   & 64.7 & 70.0   & 70.7   & 75.9  & 73.9  & 76.2  & 77.2 \\
                               & Recall & $\uparrow$   & 61.6     & 55.8 & 57.5 & 64.9   & 81.4 & 85.0   & 83.0   & 89.1  & 88.5  & 89.1  & 91.1 \\
                               & Decay  & $\downarrow$ & 2.4      & 12.6 & 2.2  & \textbf{0.0}    & 6.2  & 1.3    & 1.5    & \textbf{0.0}   & 0.6   & 7.0   & 0.9  \\ \midrule
\multirow{3}{*}{$\mathcal{F}$} & Mean   & $\uparrow$   & 50.8     & 52.3 & 55.2 & 51.1   & 66.7 & 65.9   & 65.3   & 72.1  & 68.0  & 65.3  & 72.1 \\
                               & Recall & $\uparrow$   & 60.0     & 61.0 & 51.9 & 51.6   & 77.1 & 79.2   & 73.8   & 83.4  & 80.6  & 83.4  & 83.5 \\
                               & Decay  & $\downarrow$ & 5.1      & 11.4 & 3.4  & 2.9    & 5.1  & 2.5    & 1.8    & 1.3   & 0.7   & 7.9   & \textbf{-0.2} \\ \midrule
$\mathcal{T}$                  & Mean   & $\downarrow$ & 54.8     & 65.4 & 58.2 & 60.6   & 60.3 & 71.0   & 65.4   & 66.7  & 39.0  & 39.3  & 29.1 \\ \midrule
\multicolumn{3}{c|}{Model}                           & MotAdapt & LSMO & AGS  & COSNet & AGNN & AnDiff & MATNet & GMVOS & DFNet & F2Net & Ours \\ \midrule
\multirow{3}{*}{$\mathcal{J}$} & Mean   & $\uparrow$   & 77.2     & 78.2 & 79.7 & 80.5   & 80.7 & 81.7   & 82.4   & 82.5  & 83.4  & 83.1  & \textbf{84.5} \\
                               & Recall & $\uparrow$   & 93.1     & 87.8 & 91.1 & 93.1   & 94.0 & 90.9   & \textbf{94.5}   & 94.3  & -     & 95.7  & 92.7 \\
                               & Decay  & $\downarrow$ & 5.0      & 4.1  & 1.9  & 4.4    & \textbf{0.0}  & 2.2    & 5.5    & 4.2   & -     & \textbf{0.0}   & 2.8  \\ \midrule
\multirow{3}{*}{$\mathcal{F}$} & Mean   & $\uparrow$   & 70.6     & 74.5 & 77.4 & 79.4   & 79.1 & 80.5   & 80.7   & 81.2  & 81.8  & 84.4  & \textbf{86.7} \\
                               & Recall & $\uparrow$   & 84.4     & 84.7 & 85.8 & 89.5   & 90.5 & 85.1   & 90.2   & 90.3  & -     & 92.3  & \textbf{93.3} \\
                               & Decay  & $\downarrow$ & 3.3      & 3.5  & 0.0  & 5.0    & 0.0  & 0.6    & 4.5    & 5.6   & -     & 0.8   & 0.8  \\ \midrule
$\mathcal{T}$                  & Mean   & $\downarrow$ & 27.9     & 21.2 & 26.7 & \textbf{18.4}   & 33.7 & 21.4   & 21.6   & 19.8  & 15.9  & 20.9  & 16.3 \\ \bottomrule
\end{tabular}
}
\end{table*}

\begin{table*}[t]
\caption{Quantitative results on FBMS test dataset. The best results are \textbf{bold faced}. All the other results are borrowed from \citet{lu2019see, zhou2020motion,liu2020f2net}.}
\centering
\resizebox{\linewidth}{!}{
\begin{tabular}{@{}ccc|ccccccccccccc@{}}
\toprule
\multicolumn{3}{c|}{Model}         & NLC  & FST  & FSEG & MSTP & ARP  & IET  & OBN  & PDB  & SFL  & COSNet & MATNet & F2Net & Ours          \\ \midrule
\multicolumn{1}{c}{$\mathcal{J}$} & Mean & $\uparrow$ & 44.5 & 55.5 & 68.4 & 60.8 & 59.8 & 71.9 & 73.9 & 74.0 & 56.0 & 75.6  & 76.1 & 77.5 & \textbf{77.5} \\ \bottomrule
\end{tabular}
}
\label{table2}
\end{table*}

\begin{table*}[]
\centering
\caption{Quantitative results on SegTrack-V2. The best results are \textbf{bold faced}. All the other results are borrowed from \citet{gu2020pyramid}.}
\label{table3}
\begin{tabular}{@{}ccc|cccccccc@{}}
\toprule
\multicolumn{3}{c|}{Model}         & MSTM  & STBP  & SCOM  & MBNM  & PDBM  & SSAV  & PCSA           & Ours           \\ \midrule
\multirow{3}{*}{SegTrack-V2} & $F^{max}$ & $\uparrow$ & 0.526 & 0.640 & 0.764 & 0.716 & 0.800 & 0.801 & 0.810 & \textbf{0.836} \\
 & MAE $\mathcal{M}$ & $\uparrow$   & 0.114 & 0.061 & 0.030 & 0.026 & 0.024 & 0.023 & 0.025          & \textbf{0.018} \\
 & $\mathcal{S}$    & $\downarrow$ & 0.643 & 0.735 & 0.815 & 0.809 & 0.864 & 0.851 & \textbf{0.865} & 0.860          \\ \bottomrule
\end{tabular}
\end{table*}

\subsection{Overview}

An overview of our method is depicted in Figure \ref{figure2}. Our  goal is to select the easy frame which can make mask prediction easier and to make the selected easy frame’s saliency cue better for far easier prediction. 
SOD network takes entire frames and then predicts saliency masks for every frame, as shown in the upper part of Figure \ref{figure2}. EFS selects easy frames by using images and their predicted saliency cues.
From the easy frames, the masks of the remaining frames are predicted using SVOS model. Here, we propagate masks in a bi-directional way, and that is the difference between SVOS and our UVOS approach. In SVOS, the model propagates masks from the first frame to the last frame (\textit{i.e.}, uni-direction) because the GT mask is given in the first frame, thereby the first frame is always selected as a reference frame.
Differently, the reference frame is not always the first frame in our framework. Therefore, we can propagate masks from the reference frame to both ends of the sequence. It is bi-directional. 
We call it Bi-directional Mask Prediction (BMP). We can get predicted masks for entire frames except for the selected reference frame by BMP. It is a red-filled-block in Figure \ref{figure2}. Then, we make another mask prediction from both ends to the easy frame selected. It also can be shown in Figure \ref{figure2} as a blue-filled-block, and we call it Temporal Information Updating (TIU). 

Saliency cues obtained by SOD model at first are frame-level saliency masks. Unfortunately, they cannot have any temporal information, but we use them as our beginning saliency cues. However, BMP can propagate temporal information from the easy frame to both ends. Then, both ends frame can accumulate rich temporal information. By propagating this rich temporal information via TIU from both ends to the easy frame, frame-level saliency masks are updated to video-level saliency masks. After acquiring video-level saliency masks, which have more temporal information than the beginning, EFS takes the video-level saliency masks as an input and selects new easy frames again. Then, we can get a lot better saliency masks by operating BMP and TIU. We call this iterative process Iterative Mask Prediction (IMP). At the final iteration, there is no need to further update saliency masks because all the saliency masks have enough temporal information. Therefore, we make a final prediction from selected easy frames by BMP without any more TIU.

We delineate the detail of our architecture in the remaining sections. 

\subsection{Initial Mask Prediction}

Before selecting the easy frame, we extract initial saliency masks via SOD model.
The SOD model takes entire frames and predicts foreground masks for every frame.
We adopt PFPN \citep{wang2020progressive} as our SOD network.

\subsection{Easy Frame Selector (EFS)}

Easy Frame Selector (EFS) is a module that selects easy frames among entire video frames. In EFS, the difficulty of the frame is estimated by the pair of the frame and the predicted mask. The objective of EFS is to find easy frames that show the foreground object clearly. We define the difficulty of the frame with two criteria. 1) Mask quality: if the foreground object is accurately predicted, then the frame is an easy frame.
2) Object size: if a frame has too large or small objects, there is scarce information to refer in the frame.  It may distract a prediction. It should be determined as a hard frame.

First, we use Structural-measure $\mathcal{S}$ \cite{fan2017structure}, one of the evaluation metrics for saliency masks in SOD, to evaluate the quality of our predicted masks.
$\mathcal{S}$ is a measurement that evaluates the structural similarity between predicted mask and GT mask. We need a saliency cue which reveals the object's shape well. Therefore, we design a simple CNN that estimates the $\mathcal{S}$ score from a pair of an image and its predicted mask. $\mathcal{S}$ score ranges from 0 to 1. The higher the $\mathcal{S}$ score is, the better the quality of the mask is.
Note that Mean Absolute Error (MAE) is widely used, but MAE is a pixel-wise metric. It cannot capture the structure of the object well. Experimental results on comparing $\mathcal{S}$ with MAE are presented in the ablation study section.

The network to estimate $\mathcal{S}$ is inspired by \citet{jiang2018acquisition}. \citet{jiang2018acquisition} estimates the quality of bounding boxes for objects as IoU, but we estimate the quality of a foreground mask. Also, we use a novel method to train the precise network. EFS needs to estimate the mask quality of the video frames, but the network is not trained on video datasets. Rather, we use an image dataset for training a more general and robust network. It is because video datasets have a small number of data, and they have a high correlation among the frames in the same sequence. We think that these characteristics interrupt to train robust network. In addition, the GT mask is taken as an input instead of the predicted saliency mask for training $\mathcal{S}$ in the case of perfect prediction.

Second, an object’s size is also important to estimate the difficulty of a frame. If the size of the object is too small, there only exists small information. If the size of the object is too large, the object exceed the size of the frame, and some parts of the object are missing. These frames are not proper as easy frames. Therefore, we filter out the frames with abnormal size objects through the area of the saliency mask. 

For a deeper understanding, we explain the details of EFS. Let, $X = \left\{ {{x_i}\left| {i = 0,1,...,N - 1} \right.} \right\}$ denotes a video sequence of length $N$. When $M\left(  \cdot  \right)$ denotes SOD network that detects a salient object in an image, 
\begin{equation}
    \widehat {{y_i}} = M\left( {{x_i}} \right)
    \label{equation1}
\end{equation}
where $\widehat {{y_i}}$ represents the acquired saliency mask for the image ${x_i}$ from SOD network. $F\left(  \cdot  \right)$ denotes the network that estimates $\mathcal{S}$.

The network $F\left(  \cdot  \right)$, which estimates $\mathcal{S}$, is trained with an image dataset named DUTS \citep{wang2017learning}. $\mathcal{S}$ of each image-mask pair is estimated as 
\begin{equation}
    \widehat {{S_i}} = F\left( {concat\left( {{x_i},\widehat {{y_i}}} \right)} \right).
    \label{equation2}
\end{equation}
When $S_i^{GT}$ denotes $\mathcal{S}$ score between a predicted mask $\widehat {{y_i}}$ and a GT mask ${y_i}$, the network $F\left(  \cdot  \right)$ is trained with the loss equation
\begin{equation}
    \ell  = \left| {\widehat {{S_i}} - S_i^{GT}} \right|.
    \label{equation3}
\end{equation}
A filter ${F_{size}}$ to filter out frames containing an object with abnormal size can be expressed as
\begin{equation}
    {F_{size,i}} = \left\{ {\begin{array}{*{20}{l}}
{1,}\\
{0,}
\end{array}} \right.{\rm{      }}\begin{array}{*{20}{l}}
{{\textrm{if }} t{h_{small}} < Area({{\hat y}_i}) < t{h_{large}}}\\
\textrm{otherwise}
\end{array}
\label{equation4}
\end{equation}
where $t{h_{small}}$ and $t{h_{large}}$ are thresholds of small and large object respectively and $Area( \cdot )$ is the function that can compute an area of a mask. Finally, the score to determine the difficulty of each frame can be computed as
\begin{equation}
    scor{e_i} = \hat S{}_i {F_{size,i}}.
    \label{equation5}
\end{equation}
Based on the estimated score by EFS, we select top-k easy frames as reference frames.

\subsection{Iterative Mask Prediction (IMP)}

Iterative Mask Prediction (IMP) structure iteratively selects easy frames by EFS and updates the saliency cues with temporal information. For updating part, it is composed of two parts. One is Bi-directional Mask Propagation (BMP), and the other is Temporal Information Updating (TIU).

\subsubsection{Bi-directional Mask Prediction (BMP)}

From the easy frames selected from EFS, we predict masks of the remaining frames by STM \citep{oh2019video}. All the bi-directional predictions from the easy frame to both ends of the sequence are made independently for every easy frame selected. 

\begin{figure*}[t]
\includegraphics[width=\linewidth]{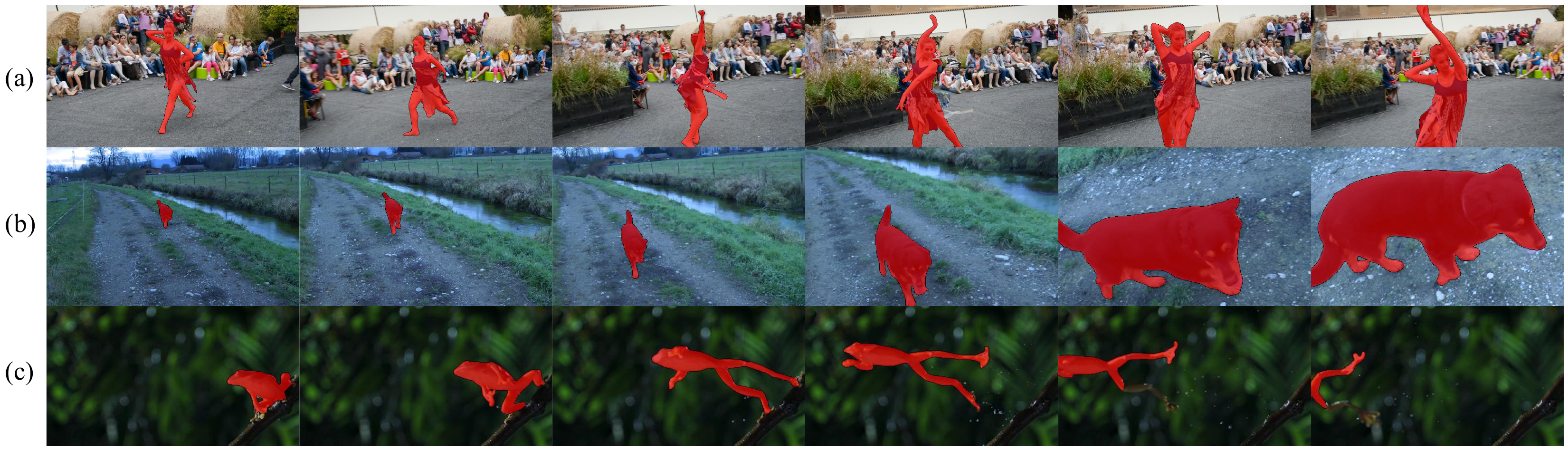}
\caption{Qualitative results from our model. (a) \textit{dance-twirl} in DAVIS16. (b) \textit{dogs02} in FBMS. (c) \textit{frog} in SegTrack-V2.}
\label{figure3}
\end{figure*}

\subsubsection{Temporal Information Updating (TIU)}

Since the saliency cues from easy frames to be used for mask prediction are brought from SOD, they do not have any temporal information. Therefore, to make saliency cues have temporal information, we update the saliency cues with TIU. Previously, BMP predicts masks from the easy frames to both ends. In TIU, saliency cues are updated by another mask prediction by STM  \citep{oh2019video} from both ends to the easy frames. At final, we aggregate entire saliency cues updated independently from each easy frame and can get more diverse temporal information for every frame. 

\subsubsection{Iterative Process}

After updating saliency cues of entire frames by TIU, EFS selects new easy frames again among updated saliency cues. We call this iterative process Iterative Mask Prediction (IMP).

\section{Experiments}

\subsection{Implementation Details}

We use PFPN \citep{wang2020progressive} as our SOD network. For mask prediction, we use STM \citep{oh2019video}. ResNet50 \citep{he2016deep} is used as a backbone network that estimates the mask’s quality in EFS. As in Equation (\ref{equation2}), an image-mask pair is fed to the network after concatenation. To match the dimension, we modify ResNet50 \citep{he2016deep}’s first convolution layer to 4 channels.
To train the network, DUTS \citep{wang2017learning}, one of SOD datasets, is used as a trainset. 

As for the iterative mechanism, our model can be seen as slow.
When we use a single easy frame and change the
number of iterations, our model consumes 0.32s per image
for two iterations, 0.44s and 0.56s for three and four iterations, respectively. Roughly, additional 0.12s is required to perform one more iteration. In constrast, when we fix the
number of iterations to three but change the number of easy
frames, our model consumes 0.44s, 0.70s, and 0.96s for two,
three, and four easy frames, respectivley. Additional 0.26s is
required to perform one more easy frame. Our method seems
quite expensive because of the iterative mechanism but it is
not true. We do not need heavy pre-processing such as optical flow and post-processing such as CRF. Based on MATNet \citep{zhou2020motion}, pre-processing and post-processing take about 0.2s and
0.5s, respectively. Thus, we believe that our model is also
competitive in terms of inference speed.

Meanwhile, if nothing is mentioned, the experiments are conducted using the number of easy frames as two and the number of iterations as four. 

\subsection{Datasets}

We conduct experiments on three benchmark datasets.

\paragraph{DAVIS16}~\citep{Perazzi2016} is a representative dataset of VOS. It is a single object VOS dataset. It is composed of 50 sequences. Among them, we evaluate our model with the validation set who has 20 sequences as other papers do. To evaluate our model, we use official metrics of DAVIS16: region similarity $\mathcal{J}$, boundary accuracy $\mathcal{F}$, and temporal stability $\mathcal{T}$.

\paragraph{FBMS}~\citep{ochs2013segmentation} is a frequently used dataset to evaluate UVOS models. FBMS is composed of 59 sequences. Among them, 30 sequences are used as a validation set. Pixel-level masks are provided as GT for some frames, not for the others. We use all the image frames for propagating richer temporal information. Then, we evaluate the performance only on the frame annotated. To evaluate on FBMS, we use region similarity $\mathcal{J}$ which is often used to evaluate on FBMS. 

\paragraph{SegTrack-V2}~\citep{li2013video} is another UVOS benchmark.
This dataset also used for Video Salient Object Detection (VSOD) task, which is a similar task to UVOS. VSOD is a task that detects salient objects in the video. Only one big difference is that salient objects can emerge from the middle frame, not the first frame, in VSOD. Fortunately, all the sequences have salient objects emerging from the first frame in SegTrack-V2, we choose this dataset to evaluate. SegTrack-V2 is composed of 14 sequences. As this dataset is originally for VSOD, we use the metrics used for VSOD task: mean absolute error $M$, max F-measure $F^{max}$, and structural measure $\mathcal{S}$. 

\subsection{Quantitative Comparison}

\subsubsection{DAVIS16}

We compare our model with other top-performing UVOS models. The results are shown in Table \ref{table1}. As can be seen in Table \ref{table1}, we achieve state-of-the-art in most metrics. Especially, we achieve very high performance in boundary accuracy $\mathcal{F}$. Due to iteratively updated temporal information, the model can find boundaries of objects more easily, which are hard to find with scarce temporal information. Compared to the previous state-of-the-art method \citep{liu2020f2net}, our model achieves gains of 1.7\% and 7.4\% on $\mathcal{J}$ Mean and $\mathcal{F}$ Mean, respectively.

\subsubsection{FBMS}

We also evaluate our model on FBMS. Since FBMS is a dataset, which is composed of images that are low-definitive and have small region of salient parts, we use four easy frames for the stable mask prediction. Also, due to the relatively short length of the sequences, we evaluate after two iterations. Compared results with other models can be shown in Table \ref{table2}. We also achieve state-of-the-art in FBMS along with \citet{liu2020f2net}.

\subsubsection{SegTrack-V2}

We evaluate our model on SegTrack-V2 and can achieve state-of-the-art in most metrics. For SegTrack-V2, two easy frames are enough to have rich temporal information and evaluate the model after two iterations. Compared to previous state-of-the-art model \citep{gu2020pyramid}, our model achieves gains of 3.2\% on $F^{max}$.

\subsection{Qualitative Results}

In Figure \ref{figure3}, boundaries of objects are clearly found. As our model predicts the foreground mask with the boundary-revealing easy frame, our model can find accurate objects’ boundaries. 

\section{Ablation Studies}

We conduct extensive ablation studies on DAVIS16 \citep{Perazzi2016} dataset.

\subsection{Number of Easy Frames}

If the number of easy frames is small, clues could be scarce. It can occur a prediction error when an object deforms a lot. On the other hand, it’s also a problem to have too many reference frames. In this case, relatively hard frames can be contained as easy frames. Then, included hard frames can deteriorate the good masks predicted by the easier frames. Therefore, it is important to set the proper number of easy frames. In Table \ref{table4}, we can find that the experiment with two easy frames has the best performance among the others. 

\begin{table}[h]
\centering
\caption{Ablation study of the number of easy frames. Using two easy frames, we could improve the performance from 84.7 to 85.5.}
\begin{tabular}{@{}c|cccc@{}}
\toprule
\# of EF & 1    & 2             & 3    & 4    \\ \midrule
$\mathcal{J}$\&$\mathcal{F}$     & 84.7 & \textbf{85.5} & 85.1 & 85.0 \\ \bottomrule
\end{tabular}
\label{table4}
\end{table}

\subsection{Number of Iterations for IMP}

\begin{figure}[h]
\centering
\includegraphics[width=\linewidth]{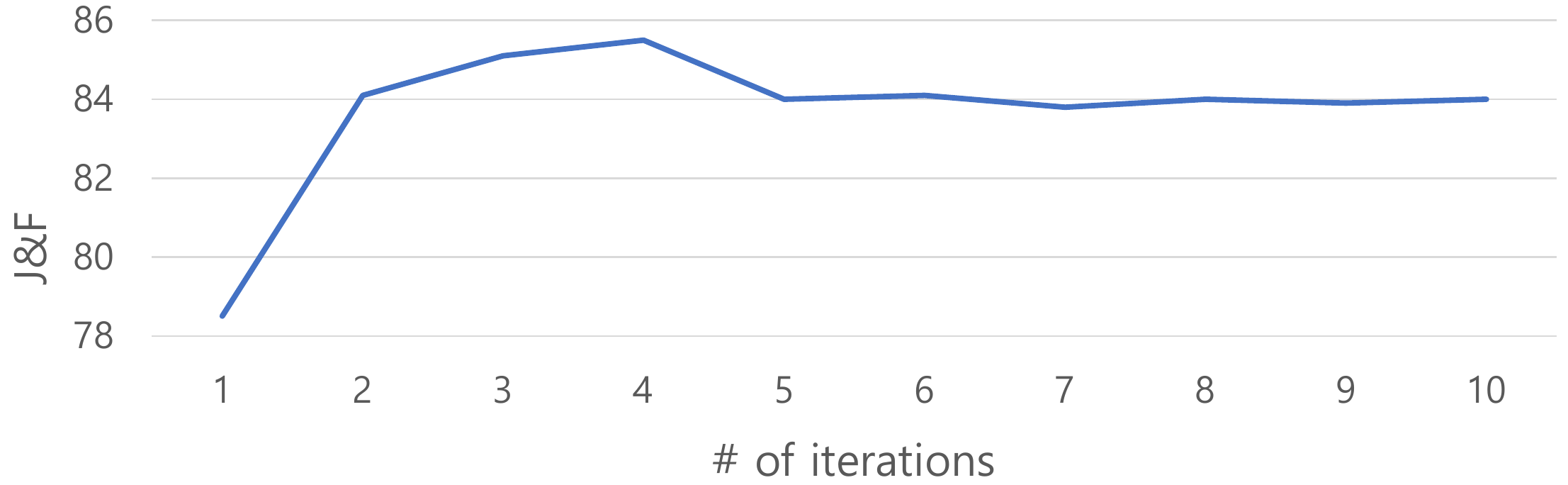}
\caption{Ablation study over the number of iterations.}
\label{figure4}
\end{figure}

In Figure \ref{figure4}, the performance increases up to four iterations. However, after four iterations, the performance is saturated with slightly lower performance. TIU updates saliency cues with richer temporal information for every iteration. Too much temporal information makes frames have too global saliency cues. It leads to a performance drop. There are almost no more updates for further iterations with TIU. Then, the performance is saturated like in Figure \ref{figure4}. Therefore, we adopt the number of iterations as four.

\subsection{Evaluation Metric for Easy Frame}

To evaluate the quality of a mask in EFS, we have two candidates to be used as an evaluation metric to estimate. The candidates are MAE and $\mathcal{S}$, which are the representative metrics for SOD. We experiment on them, and the results are shown in Table 5. Because MAE computes the mean of pixel-wise absolute error, it has a disadvantage that cannot capture the shape of an object well. Also, due to its inherent characteristic, MAE is dependent on object’s size. From the above reasons, we conclude that MAE is not proper to evaluate the mask quality. On the other hand, $\mathcal{S}$ evaluates the structural similarity between a predicted mask and a GT mask. It is proper to evaluate the quality of the mask. Therefore, we conclude to use $\mathcal{S}$ as our metric to estimate. The results in Table 5 support our decision. 

\begin{table}[h]
\centering
\caption{Ablation study on two evaluation metrics for easy frame to estimate. From the table, we can conclude that $\mathcal{S}$ is a proper metric.
}\begin{tabular}{l|cc}
\toprule
Metric & MAE    & $\mathcal{S}$\\ \midrule
$\mathcal{J}$\&$\mathcal{F}$     & 80.7 & \textbf{85.5}\\ \bottomrule
\end{tabular}
\label{table5}
\end{table}

\subsection{Effect of Temporal Information Updating (TIU)}

We ablate TIU in this study.
In Table \ref{table6}, the rightmost column is composed of experiments without TIU. They iteratively use EFS and BMP, but without TIU. As can be seen in Table \ref{table6}, the performances are dropped without TIU.
It means TIU effectively updates saliency cues with temporal information.

\begin{table}[h]
\centering
\caption{Ablation study on the effect of TIU}
\label{table6}
\begin{tabular}{@{}c|cc@{}}
\toprule
Iterations & Ours & Ours(-TIU) \\ \midrule
1          & 78.5 & 78.5       \\
2          & 84.1 & 83.7       \\
3          & 85.1 & 83.2       \\
4          & 85.5 & 83.5       \\ \bottomrule
\end{tabular}
\end{table}

\subsection{Using Other SVOS Models}

Our framework is modularized. A model used for each module can be replaceable. Above all, we experiment on replacing the SVOS model used to predict masks. Originally, we use STM \citep{oh2019video} as our SVOS model. In Table \ref{table7}, other SVOS models \citep{oh2018fast,seong2020kernelized} also can achieve good performances with our framework. Also, increasing tendency proves that our IMP is effective.

\begin{table}[h]
\centering
\caption{Ablation study on diverse SVOS models replaced. The numbers in the table are $\mathcal{J}$\&$\mathcal{F}$.}
\label{table7}
\begin{tabular}{@{}c|ccll@{}}
\toprule
Iterations & 1    & 2    & 3    & 4    \\ \midrule
RGMP       & 53.4 & 57.0 & 59.5 & 59.2 \\
KMN        & 80.0 & 83.5 & 84.2 & 84.1 \\
STM        & 78.5 & 84.1 & 85.1 & 85.5 \\ \bottomrule
\end{tabular}
\end{table}

\begin{table}[h]
\centering
\caption{Ablation study on models to predict the initial masks. Second column shows the original performances by the models. Third column shows the performance enhanced by our framework with each model. Final column shows the performance gains by our framework.}
\label{table8}
\begin{tabular}{@{}c|c|c|c@{}}
\toprule
Model  & w/o  & w/   & gain \\ \midrule
COSNet & 80.0 & 82.6 & +2.6  \\
MATNet & 81.6 & \textbf{86.8} & +5.2  \\ \bottomrule
\end{tabular}
\end{table}

\subsection{Importance of Initial Mask Prediction}

Our framework is affected by predicted initial masks. We experiment with other models to predict the initial masks. We use two UVOS models to predict: COSNet \citep{lu2019see}, and MATNet \citep{zhou2020motion}.In Table \ref{table8}, we show that our framework is effective to enhance the previous models to predict the initial masks. Especially, we achieve another state-of-the-art performance with MATNet \citep{zhou2020motion}.

\section{Conclusions}

In this paper, we address the problem of using hard frames to predict the foreground mask in UVOS task. We propose EFS to select only easy frames based on the RGB image and its predicted mask. Also, a powerful IMP structure, which iteratively selects easy frames and updates temporal information to saliency cues, achieves state-of-the-art in most metrics in three UVOS benchmark datasets. In conclusion, our proposed framework can be applied with other SVOS models. We expect that our framework can give a strong impact on an extension from SVOS models to UVOS task. 

\section{Acknowledgement}

This work was supported by Yonsei-KIST Convergence Research Program.

\bibliography{main.bib}

\begin{thebibliography}{35}
\providecommand{\natexlab}[1]{#1}

\bibitem[{Cheng et~al.(2017)Cheng, Tsai, Wang, and Yang}]{cheng2017segflow}
Cheng, J.; Tsai, Y.-H.; Wang, S.; and Yang, M.-H. 2017.
\newblock Segflow: Joint learning for video object segmentation and optical
  flow.
\newblock In \emph{Proceedings of the IEEE international conference on computer
  vision}, 686--695.

\bibitem[{Faisal et~al.(2019)Faisal, Akhter, Ali, and
  Hartley}]{faisal2019exploiting}
Faisal, M.; Akhter, I.; Ali, M.; and Hartley, R. 2019.
\newblock Exploiting geometric constraints on dense trajectories for motion
  saliency.
\newblock \emph{arXiv preprint arXiv:1909.13258}, 3(4).

\bibitem[{Fan et~al.(2017)Fan, Cheng, Liu, Li, and Borji}]{fan2017structure}
Fan, D.-P.; Cheng, M.-M.; Liu, Y.; Li, T.; and Borji, A. 2017.
\newblock Structure-measure: A new way to evaluate foreground maps.
\newblock In \emph{Proceedings of the IEEE international conference on computer
  vision}, 4548--4557.

\bibitem[{Griffin and Corso(2019)}]{griffin2019bubblenets}
Griffin, B.~A.; and Corso, J.~J. 2019.
\newblock Bubblenets: Learning to select the guidance frame in video object
  segmentation by deep sorting frames.
\newblock In \emph{Proceedings of the IEEE/CVF Conference on Computer Vision
  and Pattern Recognition}, 8914--8923.

\bibitem[{Gu et~al.(2020)Gu, Wang, Wang, Liu, Cheng, and Lu}]{gu2020pyramid}
Gu, Y.; Wang, L.; Wang, Z.; Liu, Y.; Cheng, M.-M.; and Lu, S.-P. 2020.
\newblock Pyramid Constrained Self-Attention Network for Fast Video Salient
  Object Detection.
\newblock In \emph{AAAI}, 10869--10876.

\bibitem[{He et~al.(2016)He, Zhang, Ren, and Sun}]{he2016deep}
He, K.; Zhang, X.; Ren, S.; and Sun, J. 2016.
\newblock Deep residual learning for image recognition.
\newblock In \emph{Proceedings of the IEEE conference on computer vision and
  pattern recognition}, 770--778.

\bibitem[{Huang et~al.(2019)Huang, Zheng, Huang, and Zhang}]{huang201950}
Huang, X.; Zheng, Y.; Huang, J.; and Zhang, Y.-J. 2019.
\newblock 50 fps object-level saliency detection via maximally stable region.
\newblock \emph{IEEE Transactions on Image Processing}, 29: 1384--1396.

\bibitem[{Jain, Xiong, and Grauman(2017)}]{jain2017fusionseg}
Jain, S.~D.; Xiong, B.; and Grauman, K. 2017.
\newblock Fusionseg: Learning to combine motion and appearance for fully
  automatic segmentation of generic objects in videos.
\newblock In \emph{2017 IEEE conference on computer vision and pattern
  recognition (CVPR)}, 2117--2126. IEEE.

\bibitem[{Jiang et~al.(2018)Jiang, Luo, Mao, Xiao, and
  Jiang}]{jiang2018acquisition}
Jiang, B.; Luo, R.; Mao, J.; Xiao, T.; and Jiang, Y. 2018.
\newblock Acquisition of localization confidence for accurate object detection.
\newblock In \emph{Proceedings of the European Conference on Computer Vision
  (ECCV)}, 784--799.

\bibitem[{Li et~al.(2013)Li, Kim, Humayun, Tsai, and Rehg}]{li2013video}
Li, F.; Kim, T.; Humayun, A.; Tsai, D.; and Rehg, J.~M. 2013.
\newblock Video segmentation by tracking many figure-ground segments.
\newblock In \emph{Proceedings of the IEEE International Conference on Computer
  Vision}, 2192--2199.

\bibitem[{Liu et~al.(2020)Liu, Yu, Wang, and Zhou}]{liu2020f2net}
Liu, D.; Yu, D.; Wang, C.; and Zhou, P. 2020.
\newblock F2Net: Learning to Focus on the Foreground for Unsupervised Video
  Object Segmentation.
\newblock \emph{arXiv preprint arXiv:2012.02534}.

\bibitem[{Liu, Han, and Yang(2018)}]{liu2018picanet}
Liu, N.; Han, J.; and Yang, M.-H. 2018.
\newblock Picanet: Learning pixel-wise contextual attention for saliency
  detection.
\newblock In \emph{Proceedings of the IEEE Conference on Computer Vision and
  Pattern Recognition}, 3089--3098.

\bibitem[{Lu et~al.(2020)Lu, Wang, Danelljan, Zhou, Shen, and
  Van~Gool}]{lu2020video}
Lu, X.; Wang, W.; Danelljan, M.; Zhou, T.; Shen, J.; and Van~Gool, L. 2020.
\newblock Video object segmentation with episodic graph memory networks.
\newblock In \emph{Computer Vision--ECCV 2020: 16th European Conference,
  Glasgow, UK, August 23--28, 2020, Proceedings, Part III 16}, 661--679.
  Springer.

\bibitem[{Lu et~al.(2019)Lu, Wang, Ma, Shen, Shao, and Porikli}]{lu2019see}
Lu, X.; Wang, W.; Ma, C.; Shen, J.; Shao, L.; and Porikli, F. 2019.
\newblock See more, know more: Unsupervised video object segmentation with
  co-attention siamese networks.
\newblock In \emph{Proceedings of the IEEE conference on computer vision and
  pattern recognition}, 3623--3632.

\bibitem[{Ochs, Malik, and Brox(2013)}]{ochs2013segmentation}
Ochs, P.; Malik, J.; and Brox, T. 2013.
\newblock Segmentation of moving objects by long term video analysis.
\newblock \emph{IEEE transactions on pattern analysis and machine
  intelligence}, 36(6): 1187--1200.

\bibitem[{Oh et~al.(2018)Oh, Lee, Sunkavalli, and Kim}]{oh2018fast}
Oh, S.~W.; Lee, J.-Y.; Sunkavalli, K.; and Kim, S.~J. 2018.
\newblock Fast video object segmentation by reference-guided mask propagation.
\newblock In \emph{Proceedings of the IEEE conference on computer vision and
  pattern recognition}, 7376--7385.

\bibitem[{Oh et~al.(2019)Oh, Lee, Xu, and Kim}]{oh2019video}
Oh, S.~W.; Lee, J.-Y.; Xu, N.; and Kim, S.~J. 2019.
\newblock Video object segmentation using space-time memory networks.
\newblock In \emph{Proceedings of the IEEE International Conference on Computer
  Vision}, 9226--9235.

\bibitem[{Perazzi et~al.(2016)Perazzi, Pont-Tuset, McWilliams, {Van Gool},
  Gross, and Sorkine-Hornung}]{Perazzi2016}
Perazzi, F.; Pont-Tuset, J.; McWilliams, B.; {Van Gool}, L.; Gross, M.; and
  Sorkine-Hornung, A. 2016.
\newblock A Benchmark Dataset and Evaluation Methodology for Video Object
  Segmentation.
\newblock In \emph{Computer Vision and Pattern Recognition}.

\bibitem[{Redmon et~al.(2016)Redmon, Divvala, Girshick, and
  Farhadi}]{redmon2016you}
Redmon, J.; Divvala, S.; Girshick, R.; and Farhadi, A. 2016.
\newblock You only look once: Unified, real-time object detection.
\newblock In \emph{Proceedings of the IEEE conference on computer vision and
  pattern recognition}, 779--788.

\bibitem[{Ren et~al.(2016)Ren, He, Girshick, and Sun}]{ren2016faster}
Ren, S.; He, K.; Girshick, R.; and Sun, J. 2016.
\newblock Faster R-CNN: towards real-time object detection with region proposal
  networks.
\newblock \emph{IEEE transactions on pattern analysis and machine
  intelligence}, 39(6): 1137--1149.

\bibitem[{Seong, Hyun, and Kim(2020)}]{seong2020kernelized}
Seong, H.; Hyun, J.; and Kim, E. 2020.
\newblock Kernelized memory network for video object segmentation.
\newblock In \emph{European Conference on Computer Vision}, 629--645. Springer.

\bibitem[{Siam et~al.(2019)Siam, Jiang, Lu, Petrich, Gamal, Elhoseiny, and
  Jagersand}]{siam2019video}
Siam, M.; Jiang, C.; Lu, S.; Petrich, L.; Gamal, M.; Elhoseiny, M.; and
  Jagersand, M. 2019.
\newblock Video object segmentation using teacher-student adaptation in a human
  robot interaction (hri) setting.
\newblock In \emph{2019 International Conference on Robotics and Automation
  (ICRA)}, 50--56. IEEE.

\bibitem[{Song et~al.(2018)Song, Wang, Zhao, Shen, and Lam}]{song2018pyramid}
Song, H.; Wang, W.; Zhao, S.; Shen, J.; and Lam, K.-M. 2018.
\newblock Pyramid dilated deeper convlstm for video salient object detection.
\newblock In \emph{Proceedings of the European conference on computer vision
  (ECCV)}, 715--731.

\bibitem[{Tokmakov, Alahari, and
  Schmid(2017{\natexlab{a}})}]{tokmakov2017learning2}
Tokmakov, P.; Alahari, K.; and Schmid, C. 2017{\natexlab{a}}.
\newblock Learning motion patterns in videos.
\newblock In \emph{Proceedings of the IEEE conference on computer vision and
  pattern recognition}, 3386--3394.

\bibitem[{Tokmakov, Alahari, and
  Schmid(2017{\natexlab{b}})}]{tokmakov2017learning}
Tokmakov, P.; Alahari, K.; and Schmid, C. 2017{\natexlab{b}}.
\newblock Learning video object segmentation with visual memory.
\newblock In \emph{Proceedings of the IEEE International Conference on Computer
  Vision}, 4481--4490.

\bibitem[{Tokmakov, Schmid, and Alahari(2019)}]{tokmakov2019learning}
Tokmakov, P.; Schmid, C.; and Alahari, K. 2019.
\newblock Learning to segment moving objects.
\newblock \emph{International Journal of Computer Vision}, 127(3): 282--301.

\bibitem[{Wang et~al.(2020)Wang, Chen, Zhou, Zhang, Jin, and
  Gai}]{wang2020progressive}
Wang, B.; Chen, Q.; Zhou, M.; Zhang, Z.; Jin, X.; and Gai, K. 2020.
\newblock Progressive Feature Polishing Network for Salient Object Detection.
\newblock In \emph{AAAI}, 12128--12135.

\bibitem[{Wang et~al.(2017)Wang, Lu, Wang, Feng, Wang, Yin, and
  Ruan}]{wang2017learning}
Wang, L.; Lu, H.; Wang, Y.; Feng, M.; Wang, D.; Yin, B.; and Ruan, X. 2017.
\newblock Learning to detect salient objects with image-level supervision.
\newblock In \emph{Proceedings of the IEEE Conference on Computer Vision and
  Pattern Recognition}, 136--145.

\bibitem[{Wang et~al.(2019)Wang, Song, Zhao, Shen, Zhao, Hoi, and
  Ling}]{wang2019learning}
Wang, W.; Song, H.; Zhao, S.; Shen, J.; Zhao, S.; Hoi, S.~C.; and Ling, H.
  2019.
\newblock Learning unsupervised video object segmentation through visual
  attention.
\newblock In \emph{Proceedings of the IEEE conference on computer vision and
  pattern recognition}, 3064--3074.

\bibitem[{Wu et~al.(2019)Wu, Feng, Guan, Wang, Lu, and Ding}]{wu2019mutual}
Wu, R.; Feng, M.; Guan, W.; Wang, D.; Lu, H.; and Ding, E. 2019.
\newblock A mutual learning method for salient object detection with
  intertwined multi-supervision.
\newblock In \emph{Proceedings of the IEEE/CVF Conference on Computer Vision
  and Pattern Recognition}, 8150--8159.

\bibitem[{Wu, Su, and Huang(2019)}]{wu2019stacked}
Wu, Z.; Su, L.; and Huang, Q. 2019.
\newblock Stacked cross refinement network for edge-aware salient object
  detection.
\newblock In \emph{Proceedings of the IEEE/CVF International Conference on
  Computer Vision}, 7264--7273.

\bibitem[{Yang et~al.(2019)Yang, Wang, Bertinetto, Hu, Bai, and
  Torr}]{yang2019anchor}
Yang, Z.; Wang, Q.; Bertinetto, L.; Hu, W.; Bai, S.; and Torr, P.~H. 2019.
\newblock Anchor diffusion for unsupervised video object segmentation.
\newblock In \emph{Proceedings of the IEEE international conference on computer
  vision}, 931--940.

\bibitem[{Zhen et~al.(2020)Zhen, Li, Zhou, Shang, Feng, Fang, and
  Quan}]{zhen2020learning}
Zhen, M.; Li, S.; Zhou, L.; Shang, J.; Feng, H.; Fang, T.; and Quan, L. 2020.
\newblock Learning discriminative feature with crf for unsupervised video
  object segmentation.
\newblock In \emph{European Conference on Computer Vision}, 445--462. Springer.

\bibitem[{Zhou et~al.(2020)Zhou, Wang, Zhou, Yao, Li, and
  Shao}]{zhou2020motion}
Zhou, T.; Wang, S.; Zhou, Y.; Yao, Y.; Li, J.; and Shao, L. 2020.
\newblock Motion-Attentive Transition for Zero-Shot Video Object Segmentation.
\newblock In \emph{AAAI}, volume~2, 3.

\bibitem[{Zhuo et~al.(2019)Zhuo, Cheng, Zhang, Wong, and
  Kankanhalli}]{zhuo2019unsupervised}
Zhuo, T.; Cheng, Z.; Zhang, P.; Wong, Y.; and Kankanhalli, M. 2019.
\newblock Unsupervised online video object segmentation with motion property
  understanding.
\newblock \emph{IEEE Transactions on Image Processing}, 29: 237--249.

\end{thebibliography}

\end{document}